\documentclass[16pt]{article}
\usepackage{comment}
\usepackage[paperheight=11.5in,paperwidth=9.5in]{geometry}

\usepackage{pdflscape}
\usepackage{graphicx} % import
\usepackage[utf8]{inputenc} % allow utf-8 input
\usepackage[T1]{fontenc}    % use 8-bit T1 fonts
\usepackage{hyperref}       % hyperlinks
\usepackage{url}            % simple URL typesetting
\usepackage{booktabs}       % professional-quality tables
\usepackage{amsfonts}       % blackboard math symbols
\usepackage{nicefrac}       % compact symbols for 1/2, etc.
\usepackage{microtype}      % microtypography
\usepackage{lipsum}
\usepackage{float}
\usepackage{algorithm}
\usepackage{algorithmicx}
\usepackage{algpseudocode}	 

\title{Data Analytics for Smart cities: Challenges and Promises}

\author{Farid Ghareh Mohammadi$^1$,Farzan Shenavarmasouleh$^1$, M. Hadi Amini$^2$, and Hamid R. Arabnia$^1$\\1:Department of Computer Science, Franklin College of arts and sciences,\\ University of Georgia, Athens, Georgia, 30601  \\
2: School of Computing and Information Sciences, College of Engineering and Computing, \\Florida International University, Miami, FL 33199 \\
Emails: farid.ghm@uga.edu, fs04199@uga.edu,  Amini@cs.fiu.edu, hra@uga.edu}

\begin{document}
\maketitle
\begin{abstract}
The explosion of advancements in artificial intelligence, sensor technologies, and wireless communication activates ubiquitous sensing through distributed sensors. These sensors are various domains of networks that lead us to smart systems in healthcare, transportation, environment, and other relevant branches/networks. Having collaborative interaction among the smart systems connects end-user devices to each other which enables achieving a new integrated entity called Smart Cities. The goal of this study is to provide a comprehensive survey of data analytics in smart cities. In this paper, we aim to focus on one of the smart cities important branches, namely Smart Mobility, and its positive ample impact on the smart cities decision-making process. Intelligent decision-making systems in smart mobility offer many advantages such as saving energy, relaying city traffic, and more importantly, reducing air pollution by offering real-time useful information and imperative knowledge. Making a decision in smart cities in time is challenging due to various and high dimensional factors and parameters, which are not frequently collected. In this paper, we first address current challenges in smart cities and provide an overview of potential solutions to these challenges. Then, we offer a framework of these solutions, called universal smart cities decision making,  with three main sections of data capturing, data analysis, and decision making to optimize the smart mobility within smart cities.
With this framework, we elaborate on fundamental concepts of big data, machine learning, and deep leaning algorithms that have been applied to smart cities and discuss the role of these algorithms in decision making for smart mobility in smart cities.
\end{abstract}
%In this paper, we, first, address current challenges in smart cities and provide an overview of solutions, and then to solve the aforementioned challenges we aim to offer a framework namely, universal smart cities decision making, which has three main sections: data capturing, data analysis and decision making. Lastly, we provide an abstract review of the fundamental concepts of Big Data, ML, and DL algorithms have been applied to smart cities and the essential role of the algorithms on making decision within smart cities, especially smart mobility 

% keywords can be removed
\textbf{keywords:} Smart cities, Smart Mobility, Making Decision, Artificial Intelligence, Data Science, Machine Learning, Deep Learning.

\section{Introduction}
$\noindent\diamond$\textbf{Motivation} The explosion of advancement in artificial intelligence, sensor technologies, like Internet of Things (IOT), and wireless communication activates ubiquitous sensing through distributed sensors in various domains of networks that lead us to smart systems in healthcare, transportation, environment and other relevant branches/networks. Having collaborative interaction among the smart systems connects end-user devices to each other enabling a new integrated entity called Smart Cities. 

In the last decade, the Internet of Things (IoT) devices have been connected among different independent agents and heterogeneous networks as with communication technologies \cite{hadi3}  \cite{montori2017collaborative}. The Connected high-performance sensors, and end-user devices, Internet of Things, is the trigger leveraging the networks in transitioning from urban cities towards smart sustainable cities. The goal of smart cities is to address upcoming challenges of conventional cities by offering integrated management systems with a combination of intelligent infrastructures \cite{bibri2017smart}.

Making decisions in smart cities is challenging due to high direct/indirect dimensional factors and parameters. In this paper, we aim to focus on one of the smart cities important branches, namely Smart Mobility, and its positive ample impact on smart cities decision making processes. Intelligent decision making systems in smart mobility offer many advantages, such as saving energy \cite{hadi2}, relaying city traffic \cite{an2020traffic}, and more importantly, reducing air pollution \cite{hadi1} by offering real-time useful information and imperative generated knowledge. Making an optimal decision in time in smart mobility with a wide variety of smart devices and systems is challenging. You cannot make a promising decision when your data is not frequently collected. Consequently, a training process of decision support systems still is challenging due to the lack of data \cite{mohammadi2020introduction}. In this paper, we first address current challenges in smart cities and provide an overview of potential solutions to these challenges. Then, we offer a framework of these solutions, called universal smart cities decision making,  with three main sections of data capturing, data analysis, and decision making to optimize the smart mobility within smart cities.

Interestingly, large cities are losing their own urban style and turning into smart ones. Smart cities are growing due to advanced technology, especially Artificial Intelligence(AI). The more people live in large cities, the more need to have an integrated system to cost-efficiently handle the ample growth in urbanization. The proliferation of population offers smart development challenges in such cities and enables enormous pressure on society to create innovative, smart, and sustainable environments. Therefore, today's developed cities (or smart cities) are in need of integrated smart policies and novel innovative solutions to enhance the monitoring functionality in order to facilitate urban living conditions \cite{neves2020impacts}.

Smart cities are created to enable advanced capabilities, such as sustainable energy systems and electrified transportation networks, and interact with information and communication technologies (ICTs) to enhance the efficiency of the cities being generated \cite{hadi4, park2021emerging}. An example of responding to new changes for smart cities may be seen in the progress being made in smart healthcare systems for emergency care cases. To enable Hospitals to monitor and control their patients and let the specialists offer better solutions to their patients, they need to use a smart healthcare system \cite{hossain2019smart} \cite{ pacheco2019smart}.

To develop the smart healthcare system, we need to use healthcare networks that are the inter and intra connection among the healthcare components like IOT devices and sensors to enhance the process of monitoring and consequent services for patients \cite{ellaji2020efficient, hadi2}. The performance of such a system heavily depends on the quality of this network communication or online synchronization with other associated networks (other smart systems in smart cities) that actively contribute to the service operation in a positive way. For example, a healthcare network may  take advantage of networks' resources, like energy, in case the system fails to run an operation due to lack of enough power supply. In this situation, the best solution is connecting this network with other networks for the benefit of both patients and their health and service management reliability \cite{hadi2}.

\begin{figure}
     \centering
     \includegraphics{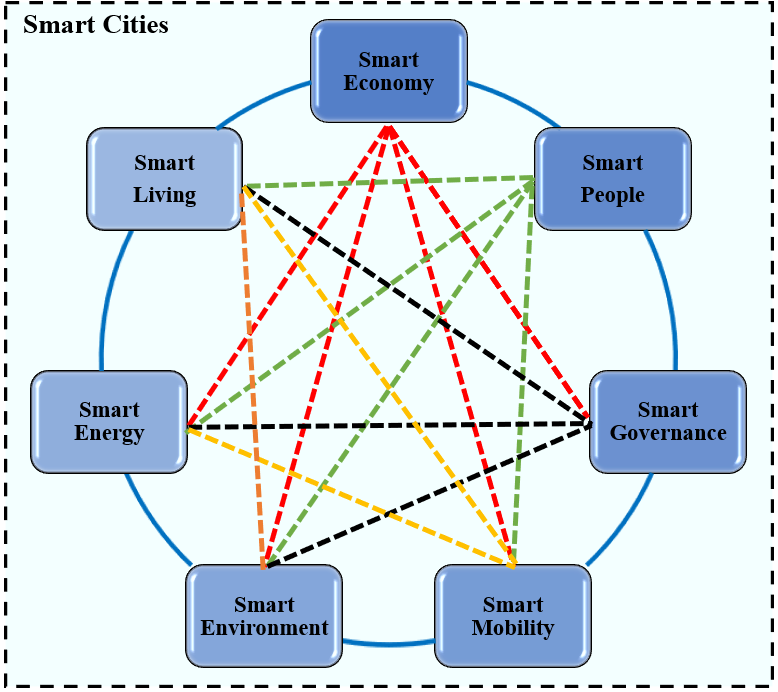}
     \caption{Smart cities main features}
     \label{fig:smart_Cities}
 \end{figure}

%Making Cities and turning them into Smart cities require some features representing them carefully.
%Identifying several features critical to the conversion of urban centers into smart cities.
In the processes of making and turning cities into smart cities, we can identify several representative features. According to a research study \cite{tran2019application}, the scientists investigated a period of 10 years of work from 2008 to 2018 discovering that smart cities might have some features in common. The most interesting features \ref{fig:smart_Cities} \cite{tran2019application, de2020development, hadi4} are smart economy, smart people, smart governance, smart mobility, smart environment, smart energy, and smart living that are shown in figure \ref{fig:smart_Cities}. As shown in the figure \ref{fig:smart_Cities}, although each feature represents the importance of itself to smart cities, they all are inter-related to each other. Each feature has direct or indirect impact on others. Furthermore, proposed decision-making solutions used in smart cities are Multi-criteria decision-making (MCDM), Mathematical Programming (MP), Artificial Intelligence (AI), and Integrated Method (IM). In this study, we aim to discover AI solutions, particularly Machine learning and deep learning approaches rather than others, for smart mobility. 

\section{Role of Machine Learning in Smart Cities}  
Current advanced technologies in sensors and Internet of Things (IoT) devices make it essential in leveraging Artificial Intelligence, particularly Machine Learning, to model the data for further application. \cite{boulos2019overview, al2020intelligence}. The IoT devices are considered as the most important and unavoidable parts of smart cities. These devices provide a huge amount of data depending on which applications are going to be used, such as healthcare and transportation applications \cite{ullah2020applications}.

Internet of Things technologies have proliferated in many fields, such as smart healthcare \cite{ellaji2020efficient, ellaji2020efficient, manikandan2020hash} and smart home systems like Alexa \cite{khatri2018advancing, rak2020systematic}, specifically in urban cities, turning them into smart cities. Thus, the huge usage of Internet of Things technologies plays a pivot role in generating big data which requires solutions to analyze and keep track of smart cities activities. This big data analysis \cite{elhoseny2018framework} provides invaluable knowledge in order to integrate all smart cities' sources like IOT and networks. As smart cities and their data grow, this analysis process may become challenging for future decision-making.
%As smart cities grow, this analysis may become challenging towards their future decision-making Problems. 
In the next section, we address this problems and discuss possible solutions have been proposed thus far.
Researchers in \cite{ju2018citizen} proposed a new solution to leverage citizen-centered big data analysis to apply to smart cities. Their solution is to determine a path for implementation of citizen-centered big data analysis for the sake of decision-making. This solution's main goal is to provide imperative perspectives: data-analysis algorithms and urban governance issues \cite{ju2018citizen}.
 
Furthermore, researchers in \cite{bhattacharya2020review} provided several deep learning applications in smart cities,
such as smart governance, smart urban modeling, smart education, smart transportation,intelligent infrastructures, and smart health solutions. Additionally, 
the challenges of using deep learning towards smart cities are also addressed. However, still the problem of decision making in smart cities remains challenging. In the next section, we address the problems and highlight the possible solutions.

 \section{Smart Cities Data Analytics Framework}
 There are plenty of research studies accomplished in smart cities like \cite{kumar2018intelligent}, which is an intelligent decision computing solution for crowd monitoring. In this section, we dig into smart cities applications, and the foundation of techniques are developed upon by establishing a universal smart cities data analytics framework, which is elaborately depicted in figure \ref{fig:data_analytics}. This framework has three main sections: Data Capturing, Data Analysis, and Decision Making. We elaborate on each in separate sections.
 \subsection{Data Capturing}
\begin{figure}
     \centering
     \includegraphics[height=3.5in]{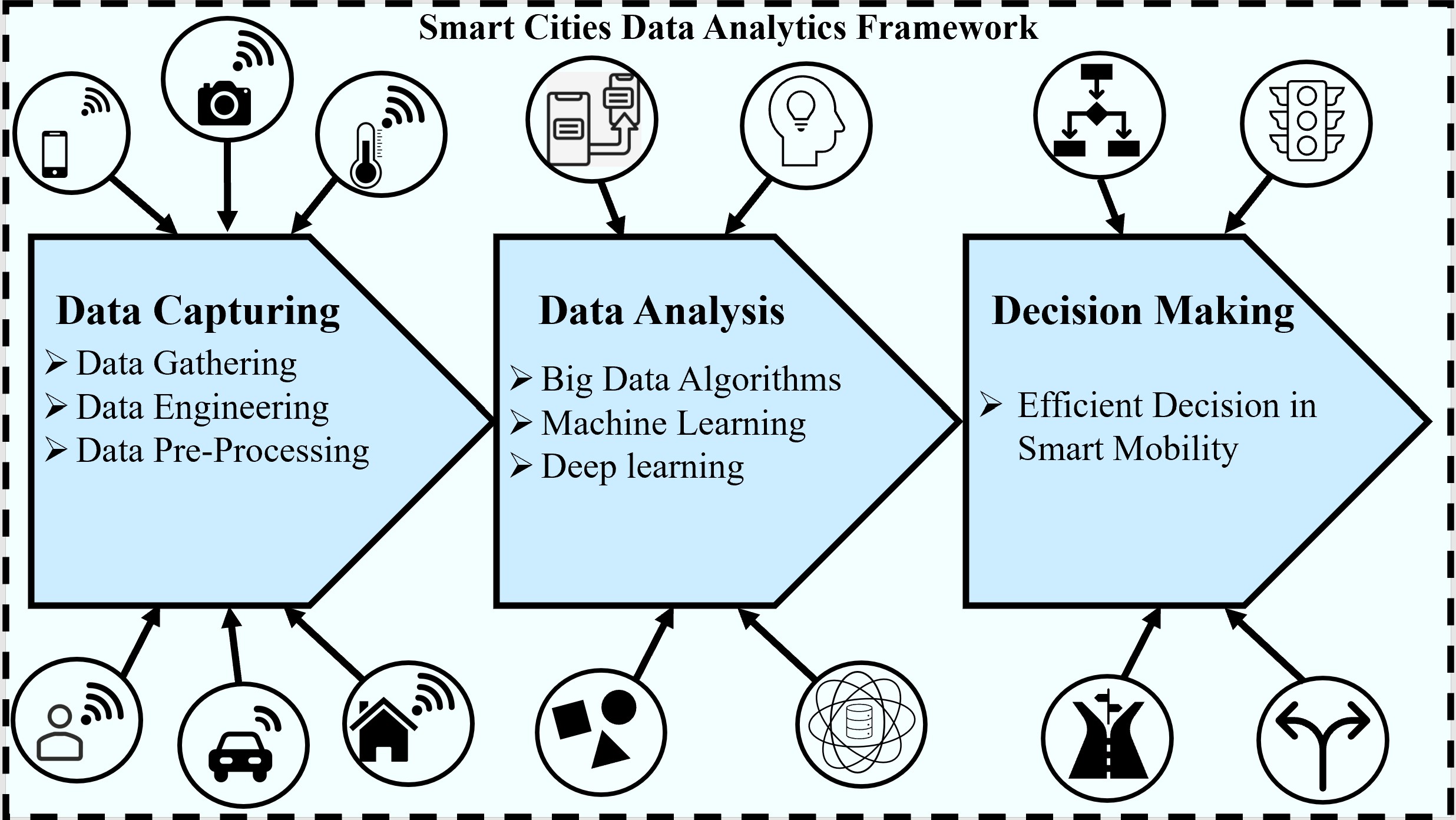}
     \caption{Smart cities data analytics framework}
     \label{fig:data_analytics}
 \end{figure}

 Smart cities have been engulfed with too much data which requires the management department to control and monitor the cities, but this department is cost and time inefficient. Hopefully, due to domains (features of smart cities), these data ,which are grouped automatically into domains, create particular big data for each domain separately. IOT devices used to gather data for healthcare are completely different from the ones that are developed specially for traffic control (i.e that is why we need to categorize the data into groups of domains to create certain technical big data for each domain).
 According to figure \ref{fig:data_analytics}, we provide 6 different samples of sensors and IOT devices, such as smart phones, smart cameras, smart thermometers, smart users with sensors, smart cars, and smart houses. The variations of sensors enable a data center to receive different types, ranges, and values of data from objects.  Here, we highlight the current upcoming challenges with associate solutions, respectively.
 
 $\noindent \diamond$ \textbf{Challenges:} The process of data gathering in smart cities has been enhanced yet remains challenging due to lack of enough equipment in every place we aim to monitor and make decisions for that region, like relaying traffic information. No matter our focus, whether in smart healthcare mobility or other areas of interest, IOT devices and other sensors may still fail to capture all data correctly or may miss data entirely due to their limited storage and inefficient time-scales.
%Based on our focus whether in smart healthcare or smart mobility or the rest of smart cities' areas of interest, having associated IOT devices and other sensors in the focused spot fail to capture all data correctly or miss some data due to their limitations such as lasting time and limited storage.
 
 In addition to that, data have proliferated significantly and are produced from heterogeneous sources. Therefore, the types of data vary, from video and images to digits or strings, and need particular procedures to convert all of the data into single unit measurements. These measurements enable us to run machine learning algorithms and other deep learning algorithms on the data readily to make optimum decisions.
 
 $\noindent \diamond$ \textbf{Solutions:} Pre-processing plays a pivotal role in managing missing information and values within the generated dataset. There are some basic and advanced approaches \cite{usman2019survey} to handle the missing values, and also, there are tools and techniques to select important and relevant features like IFAB \cite{mohammadi2014image} using Artificial Bee colony to get rid of irrelevant features and ensure the genuineness and reproducibility of the results \cite{farzanstat2019}.

 To handle the heterogeneous problem, Data Engineering \cite{elhoseny2018framework} is responsible for managing and analyzing the input data and adding labels to data left unlabeled. This requires experts and time, thus it is not cost and time efficient. Therefore, leveraging Big Data algorithms helps to tune and  analyze the data properly.
 \subsection{Data Analysis}
 The smart cities promises lead us to an ample proliferation and generation in data from all aspects of the domains and branches. Therefore, such huge amounts of data are at the core of the services generated by the IoT technologies \cite{hashem2016role}. This section of the framework, Data Analysis, is imperative because its results lead us to make proper decisions. If this process is not accomplished, the decision made will not be efficient. Thus, a large number of research studies have enhanced the process and yielded better results. In the early era of smart cities, there were only limited data generated everyday due to the lack of sensors.`Therefore, typical machine learning algorithms were sufficient for data analysis to make a model that can handle the situation and provide enough information to make a decision. %It does mean that those algorithms did not face any challenges, which are discussed later this section.
 However, thanks to technologies, the number of sensors and IOT objects have proliferated and thus we have huge amounts of data that require big data algorithms like  Spark and Hadoop to handle the data. \cite{mohammadi2020Covid}
 
 Additionally, due to the huge quantity of data, researchers used deep learning algorithms, especially Transfer learning and Meta-learning \cite{mohammadi2020introduction} and some other famous machine learning techniques to learn within Reinforcement Learning like Q-learning \cite{chakole2021q, liao2020fast,fan2020theoretical} for generating smart systems \cite{boussakssou2020towards, joo2020traffic}.
 
\subsubsection{Big Data Algorithms and challenges}
Due to the Big Data revolution, the enormous volume of high performance computations are unavoidable in such smart cities. Thus, big data algorithms are getting one of the important functioning pieces of smart cities. 

$\noindent \diamond$ \textbf{Transportation System:} Transportation systems are generated using advanced technologies where applications of big data vary and are important \cite{wang2021review}. Scenarios are established in the research studies to offer the following options to smart people: suggesting the best travel time for any given  trip, providing real time
traffic information, and predicting movement patterns according to personality (daily routing path) or spatio-temporal
routines, enhancing crash analysis, planning bus routes, improving taxi dispatch, and optimising traffic time during big occasions \cite{wang2021review}. For example, Zhu \emph{et al} in \cite{zhu2018big}, developed an approach to use two algorithms: Bayesian inference and Random forest to execute in real-time to predict the probability of crash occurrence to decrease crash risks in smart cities. Moreover, researchers \cite{xiong2014novel}  have established a combination of supervised and regression algorithms, such as multivariate adaptive regression splines, regression trees, and logistic regression, to study motor vehicle accident injury behaviour.

$\noindent \diamond$ \textbf{Urban Governance}
Learning from human mobility patterns discloses the movement and trend of a large population, the most popular applications of which are in domains like crime prediction \cite{wang2021review}, disaster evacuation, big events management, and safety estimation. To that end, knowledge about big events derived from bus trajectories, including event start time and end time, can be beneficial for the event planning and management department \cite{mazimpaka2017they}.

\subsubsection{Machine learning process and challenges}
The learning process based on given input where data are not sufficient does not produce reliable and robust results \cite{mohammadi2020introduction}. Machine learning algorithms get stuck in local minimum or maximum easily due to a large amount of data which leads to a problem called over-fitting \cite{mohammadi2020introduction}. Furthermore, machine learning algorithms fail to learn positions and states (classes) when the number occurrence in the data is far less than others. For instance, within the smart healthcare domain, when capturing data, the probability of having all desired classes looks low and the number of rare diseases would be insufficient for learning.
  
$\noindent \diamond$ \textbf{Solutions:} In paper \cite{mohammadi2019promises}, researchers established an approach called  Meta-Sense, i.e. learn to sense rather than sense to learn. This approach took advantage of one of the important data analytic promises called zero-shot learning (ZSL) \cite{wei2020lifelong, mohammadi2020introduction} . There are many applications of ZSL in computer vision \cite{mohammadi2019parameter}  and motion detection in healthcare application \cite{wijekoon2018zero} %(Add CSCE'20 paper here). 
Furthermore, researchers in \cite{yu2020zero} proposed a solution based on ZSL called Zero-Virus providing a deep understanding for an intelligent transportation system to generate the best rout for drivers. Zero-virus does not need any vehicle-tracklets annotation, thus it is the most volatile real-world traffic scheme. 

$\noindent \diamond$ \textbf{Machine learning statistic records:} Iskandaryan \emph{et al} in \cite{hadi1} investigated a survey to study sensor data  and analyze the impact of air quality in Smart Cities  using supervised machine learning algorithms. They established four important and popular categories of algorithms associated with the number of research studies: first, the most popular ML algorithm, neural network (NN), second, Logistic regression, third, ensemble algorithm, followed by others. Furthermore, the results in \cite{hadi1} demonstrated that the number of publications associated with machine learning algorithms with the application of smart cities, particularly smart environment by predicting and preventing the air pollution risks, have increased progressively. Additionally, researchers in \cite{hadi1} evaluated the number of metrics that have been used in publications to examine the performance of each algorithm. The two most popular metrics are Root Mean Square Error  (RMSE) and  Mean Absolute Error (MAE).

\subsubsection{Deep learning process and challenges}
Feasible smart cities are established by technology driven foundations and their initiatives are on different branches and domains, in which each of them may requires systems with high-performance computing resources and technologies. Such systems have pros and cons such as saving energy and reducing the air pollution and reducing diagnose time, but (writing something bad). One of the most popular technologies used to tackle such huge amounts of data is Deep Learning (DL). DL is a special algorithm within machine learning  and is efficiently used to obtain required knowledge from the input data, extracting the patterns which govern the whole data and also classify them. Several research studies are successfully applying DL on smart cities \cite{bhattacharya2020review}, such as urban modeling for smart cities, intelligent infrastructure for smart cities, and smart urban governance. Here, we aim to focus on a particular application of deep learning which is smart mobility and transportation.

AI definitely pushed all the science a step forward by making the systems and processes of scientific inquiry as smart as possible, for example autonomous transportation systems \cite{bhattacharya2020review} in smart mobility. Making a decision about whether the object seen is a human being or not is challenging \cite{pandey2018object}. Object detection is one of the challenging issues in smart mobility that surely boosts and facilitates automation in transportation systems. Consequently, this positively enhances and improves  smart mobility in Smart Cities. Researchers in \cite{pandey2018object} explored analysis of decent object detection solutions like deep learning \cite{Asali2020DeepMSRFAN}. The scientists leveraged a well-known object detection system, namely YOLO (You Only Look Once), which was developed earlier by Redmon and \emph{et al} in \cite{redmon2016you} and assessed its performance on real-time data.

\subsubsection{Learning process and emerging new type of data problems} 
In this section, we address possible challenges and solutions in the world of data analysis. The first and foremost problem that researchers tackle is lack of data for rare classes within the dataset that are used to make a model. The less number of samples we have in the dataset, the higher chance of ignoring that sample while we train and make the model. To handle this problem, Metalearning has come to play an essential role to make a model only using few samples. It has three important promises: zero shot learning, one shot learning, and few shot learning \cite{mohammadi2020introduction}.
Zero shot learning is a certain type of Metalearning when a training dataset does not have any samples for classes, and we still can predict them during a test process. For instance, a research work \cite{yu2020zero} was conducted to not use any annotation for processing vehicles-tracklets. This study established a route understanding system based on zero-shot theory for intelligent transportation, namely zero-virus, which obtains high effectiveness with zero samples of annotation of vehicle tracklets. Further, another research work \cite{mohammadi2019promises} established a new technique, namely Metasense, which is the process of learning to sense rather than sensing to learn. This process takes advantage of the lack of samples of classes by learning from learning rather than learning from samples.

The process of non-annotation helps to work with data that lacks classes because typical machine learning algorithms fail to detect non-annotation classes. Furthermore, sensing to learn helps new algorithms predict information that is completely lacking from the dataset itself. These advancements lead us closer to (or are part of) zero shot learning, which is critical for the advancement of Smart Cities.

The second promise is one shot learning, in which each epoch in a training phase has only one sample per each class which is taken by a deep learning algorithm or a combination of neural networks  \cite{mohammadi2020introduction}.

The third promises but not the least one, is few (k) shot  learning, in which each epoch in a training phase has only few (k) samples per each class which are taken by a deep learning algorithm or a combination of neural networks  \cite{mohammadi2020introduction}.

\subsection{Decision Making Problems in smart cities}  
Decision making problems are becoming challenging issueS in smart cities where not only the problem itself but also other relevant problems in other aspects of smart cities need to be analyzed.
%where are needed to analyze not only the problem itself, but also relevant problems and all aspect of smart cities.
Additionally, decision makers must depict the consequence of the decision they are going to make. Thus, decision making systems are needed  in smart cities in which the systems take care of all issues within the connected networks and only some limited information is taken which would be enough make a optimum decision. In this section, we highlight the challenging decision making problems and solutions.

\subsubsection{Traffic decision making system}
  Traffic decision making system, as known as intelligent transportation system (ITS), aims to detect traffic flow within a smart city and offer optimal solutions using proper big data analytics \cite{safari2020fast}.  Assessing and analysing this big data plays a pivot role in decision making systems that makes the process time- and cost-efficient \cite{zhu2018big}. Researchers addressed some case studies withing traffic decision making systems, such as road traffic accidents analysis \cite{golob2003relationships, xu2015mining}, public transportation management and control, and road traffic flow prediction. 
  
$\noindent \diamond$ \textbf{Challenges:}
In these such systems, users feed data into them by gathering data from heterogeneous resources, such as high-performance IOT devices, video detectors, and GPS. The systems must use big data analysis approaches to evaluate the data online \cite{mohamed2014real} in order to provide efficient and intimate knowledge for decision making. To be more certain about the amount of data these systems handle, consider Petabyte level data which are beyond the utmost of  traditional machine learning analytic abilities \cite{mohamed2014real}. This is due to two important issues: these traditional data processing algorithms are not yet developed for online and real-time monitoring systems, and additionally, they fail to learn from these data due to disorganized and nonstandard structures. 
  
$\noindent \diamond$ \textbf{Solutions:}
Smart cities, particularly in this domain (i.e one of their important features, smart mobility, discussed in figure \ref{fig:smart_Cities}), need optimum traffic decision making systems which address the aforementioned challenges and provide efficient solutions to them. One of the solutions discussed in \cite{mohamed2014real} is the use of big data analytics approaches. They solved the challenges by providing enough data storage,  analysis and management tools. The most important and promising frameworks, libraries or technologies are used to analyze the big data, such as Hadoop and Spark, but not limited to them. For instance, researchers in \cite{park2020pacc} established a time-efficient and scalable distributed technique, namely Partition-Aware Connected Components (PACC), for connected component computation that relies on main approaches: edge filtering, two-step processing of partitioning and computation, and sketching. PACC outperforms MapReduce and Spark frameworks and is time-efficient on real-world graphs \cite{park2020pacc}. 

Additionally, Big Data analysis algorithms facilitate and boost the process of handling large amount of data to provide enough information for making decisions. Traffic decision making systems using Big Data algorithms and technologies help associate offices and the people in charge to get to learn  drivers journey patterns within the transportation network where reports whole networks trends or better understanding of similar drivers \cite{zhu2018big} . Considering this feature, the systems provide the best path to drivers to reach their destination through the minimum time possible. Furthermore, the systems help the city to relay the traffic by controlling traffic light. The best timing to make which light should be on or off and for what period.

Furthermore, Traffic decision making systems predicts the the probability of traffic accident occurrence using Big Data algorithms \cite{zhu2018big}. This requires to have smart healhthcare systems, which is nominated as one of smart cities features discussed in figure \ref{fig:smart_Cities}, to help emergency centers to facilitate the process of emergency rescue.

\subsubsection{ Safe and Smart Environment} Researchers in \cite{an2020traffic} leveraged deep learning algorithms and advanced communication technologies to link vehicles, roads and drivers in order to facilitate and enhance various traffic‐related tasks and improve air pollution \cite{hadi1}. The scientists focused on different initiatives with the goal of creating a safe and smart environment and transportation. Further, Zhu \emph{et al} in \cite{zhu2018big}, developed an approach to use two algorithms: Bayesian inference and Random forest to execute real-time tp predict the probability of crash occurrence to decrease crash risks in smart cities. Moreover, researchers \cite{xiong2014novel}  have established a combination of supervised and regression algorithms such as multivariate adaptive regression splines, and classification and regression trees, logistic regression and  to experiment analytical study using of a dataset of motor vehicle accident injury.

To have a safe environment, air pollution prediction in smart cities plays imperative role. There have been a significant amount of work have tied to enhance the prediction using several machine learning algorithms. The usage of machine learning algorithms to make environment safe has increased consistently, stating that how important this prediction would be for smart cities \cite{hadi1}. We examined the most relevant research studies \cite{hadi1, hadi2, ju2018citizen, xiong2014novel} but not limited to these, applying different evaluation based on several metrics. The evaluation of those research lead us to following common observations: first, the rate of applying of the advanced (deep learning) machine learning algorithms have proliferated rather than typical machine learning algorithms; second, among the prediction elements for air pollution prediction, PM2.5 is considered as the most popular element; third, the data used for air pollution prediction had already generated hourly rather than daily; and final observation, efficient prediction occurs when air-quality captured data merged with other relevant data of other networks like healthcare network within the smart cities.

\section{Conclusion}
We highlight smart cities’ research branches and technology advancement regarding different complex domains. We propose a solution namely, universal smart cities decision making, which has three main section: data capturing, data analysis and making decision. We provide an abstract review of the fundamental concepts of Big Data, ML, and DL algorithms have been applied to smart cities. We explore the essential role of the aforementioned algorithms on making decision within smart cities. The goal of this study is to provide a comprehensive survey of data analytics in smart cities, more specifically, the role of Big Data algorithms and other advanced technologies like ML and DL for making decision in smart mobility within smart cities.

\bibliographystyle{unsrt}  
\bibliography{smartCity}  %%% Remove comment to use the external .bib file (using bibtex).
%%% and comment out the ``thebibliography'' section.

\end{document}